\documentclass[journal]{IEEEtran}
\IEEEoverridecommandlockouts
\usepackage{cite}
\usepackage{amsmath,amssymb,amsfonts}
\usepackage{algorithmic}
\usepackage{graphicx}
\usepackage{textcomp}
\usepackage{xcolor}
\usepackage{caption}
\usepackage{amssymb}
\usepackage{booktabs}
\usepackage{graphicx}
\usepackage{caption}
\usepackage{float}
\def\BibTeX{{\rm B\kern-.05em{\sc i\kern-.025em b}\kern-.08em
    T\kern-.1667em\lower.7ex\hbox{E}\kern-.125emX}}
\begin{document}

\title{Exploring Facial Biomarkers for Detecting Depression through Temporal Analysis of Action Units\\}

\author{
Aditya Parikh$^{1}$,
Misha Sadeghi$^{1}$,
Robert Richer$^{1}$,
Lydia Helene Rupp$^{2}$,
Lena Schindler-Gmelch$^{2}$,
Marie Keinert$^{2}$,
Malin Hager$^{2}$,
Klara Capito$^{2}$,
Farnaz Rahimi$^{1}$,
Bernhard Egger$^{3}$,
Matthias Berking$^{2}$,
Bjoern M. Eskofier$^{1,4}$\\[1ex]
\small
$^{1}$Machine Learning and Data Analytics Lab, Department Artificial Intelligence in Biomedical Engineering (AIBE), Friedrich-Alexander-Universität Erlangen-Nürnberg (FAU), Erlangen, Germany\\
\small
$^{2}$Department of Clinical Psychology and Psychotherapy, Friedrich-Alexander-Universität Erlangen-Nürnberg (FAU), Erlangen, Germany\\
\small
$^{3}$Chair of Visual Computing, Department of Computer Science, Friedrich-Alexander-Universität Erlangen-Nürnberg (FAU), Erlangen, Germany\\
\small
$^{4}$Translational Digital Health Group, Institute of AI for Health, Helmholtz Zentrum München – German Research Center for Environmental Health, 85764 Neuherberg, Germany
}

\maketitle

\begin{abstract}

Depression is characterized by persistent sadness and loss of interest, significantly impairing daily functioning and, now a widespread mental disorder. Traditional diagnostic methods rely on subjective assessments, necessitating objective approaches for accurate diagnosis. Our study investigates the use of facial action units (AUs) and emotions as biomarkers for depression. We analyzed facial expressions from video data of participants classified with or without depression. Our methodology involved detailed feature extraction, mean intensity comparisons of key AUs, and the application of time series classification models. Furthermore, we employed Principal Component Analysis (PCA) and various clustering algorithms to explore the variability in emotional expression patterns. Results indicate significant differences in the intensities of AUs associated with sadness and happiness between the groups, highlighting the potential of facial analysis in depression assessment.

\end{abstract}


\section{Introduction}

The World Health Organization (WHO) estimates that depression affects millions of people worldwide, making it one of the most common mental health disorders. Depression is a condition that considerably reduces everyday operations and quality of life. It is characterized by continuous sorrow, lack of interest in activities, and a variety of mental and physical difficulties \cite{who_report2}. A timely and accurate diagnosis is essential for managing and treating conditions effectually. Traditional diagnostic techniques frequently depend on subjective self-reported questionnaires and clinical interviews like the Beck Depression Inventory (BDI) \cite{beck}, the Hamilton Depression Rating Scale (HDRS) \cite{dep_rating}, and the Patient Health Questionnaire (PHQ-8 and PHQ-9) \cite{phq}. Several factors could influence these evaluations.

More objective, trustworthy, and quantitative diagnostic tools are needed to supplement these established techniques. One promising approach involves the analysis of facial expressions and emotions, as these can provide objective and non-invasive indicators of mental health status \cite{facs_depression}. Previous studies have demonstrated that certain facial expressions and feelings are more prevalent in individuals with depression \cite{sadness_happiness}. 

Using facial expression analysis to find biomarkers linked to depression is one potential method. Facial expressions, which are governed by the Facial Action Coding System (FACS)'s definition of facial action units (AUs), are widely used markers of emotional states and can reveal important information about a person's mental state \cite{facs}. The FACS, proposed by Ekman and Friesen \cite{facs}, provides an exhaustive framework for categorizing facial movements associated with specific emotions. This system allows for the detailed analysis of facial action units, which are the fundamental actions of individual facial muscles \cite{facs_depression}. Additionally, people with depression often exhibit a higher prevalence of sadness and lower frequencies of happiness compared to healthy individuals. However, there is a need for more refined and quantitative analyses to better understand the relationship between AUs, emotions, and depression.

This study aims to investigate the potential of facial AUs as biomarkers for depression through a detailed temporal analysis. Temporal analysis refers to examining how AUs change and interact over time. Instead of merely looking at static facial expressions, we analyze the dynamic sequences and patterns of these expressions to understand their temporal characteristics. By examining the dynamic patterns of facial expressions, we seek to identify specific AUs that are indicative of depression and to develop predictive models that can accurately distinguish between individuals with and without depression. Our study contributes to a thorough examination of the temporal dynamics of facial expressions in depression, adding to the expanding corpus of research on objective mental health screening instruments. The results of this research could improve the precision of diagnoses and facilitate the creation of automated, non-invasive depression screening instruments, which would eventually improve patient outcomes and accelerate intervention times. 

\section{Related Work} \label{section:studies}

The relationship between facial expressions, depression, and emotions has been studied broadly, providing significant insights into the possible applications of face analysis in mental health evaluations. This section reviews key studies that have identified dominant facial AUs and emotions in people with depression, as well as the techniques used to conduct these analyses.

Jones et al. (2018) \cite{jones2018} conducted an extensive study on the facial expressions of depressed and healthy individuals using the FACS. Their analysis revealed that depressed patients exhibited higher frequencies of AU1 (inner brow raiser), AU4 (brow lowerer), and AU15 (lip corner depressor), which are commonly associated with sadness and distress. Also, the study found that these patients displayed lower frequencies of AU12 (lip corner puller), which is related to expressions of happiness.

Li et al. (2020) \cite{r1} explored the use of machine learning models to classify depression based on facial expression data. They extracted a wide range of AUs and employed Principal Component Analysis (PCA) to reduce dimensionality before applying support vector machines (SVM) for classification. Their findings showed that AUs associated with sadness (e.g., AU1, AU4) and reduced expressions of happiness (e.g., AU12, AU25 - lips part) were significant predictors of depression. The study demonstrated the feasibility of using automated facial analysis and machine learning for depression detection. 

Zhang et al. (2022) \cite{r2} proposed a hybrid model combining convolutional neural networks (CNNs) and recurrent neural networks (RNNs) to capture both spatial and temporal features of facial expressions in depressed patients. By leveraging long short-term memory (LSTM) networks, their model effectively captured the temporal dynamics of facial expressions, leading to a significant improvement in depression classification performance. According to their research, temporal information must be taken into account to reliably identify depression from facial cues.

A more recent study by Wang et al. (2023) \cite{r3} utilized a multi-modal deep learning approach, integrating facial expression analysis with audio and textual data to detect depression. Their multi-modal model showed superior performance compared to traditional models, highlighting the importance of combining different types of behavioral data for comprehensive depression assessment. This study provided plausible evidence for the potential of multi-modal deep learning frameworks in mental health diagnostics.

These studies collectively highlight the importance of specific facial AUs and emotions in identifying depression. They provide a foundation for our research, which further investigates the dominance of certain AUs and emotions in depression versus participants without depression. 

\section{Methodology}\label{section:Methodology}

\subsection{Data Collection}

Data collection was part of the EMPKINS Subproject D02 - Empatho-Kinaesthetic Sensor Technology for Biofeedback in Depressed Patients\cite{do2}. The goal of this sub-project is to train smartphone-based reappraisal training for psychological assessment using facial expressions as biomarkers. The project aims to understand the relationship between cognition, facial expressions, and affect as underlying mechanisms for the development and maintenance of depression, addressing the lack of empirical studies on facial expressions and the need for quantifying and modifying them.

Each participant was provided with a mobile application designed for the study. Participants were asked questions and expressed emotions were recorded using their smartphones and a camera setup. This process was conducted in the presence of a psychologist, encompassing multiple phases to capture a diverse range of emotional expressions. 

\subsection{Feature Engineering}

The recorded video data for this study were processed and exported to CSV files using the OpenDBM tool \cite{r4}. These CSV files contained detailed frame-wise information about facial expressivity and AUs. Specifically, OpenDBM provided features such as the intensity and presence of individual AUs, as well as composite metrics for overall facial expressivity. The videos were recorded at a frame rate of 30 frames per second (fps).

To handle the large volume of data efficiently, a Python script was developed to batch process the videos on a high-performance computing (HPC) cluster. This script automated the extraction of relevant features and ensured consistent processing across all video files.

\subsubsection{Data Preprocessing}

The exported files included two key files for each video as CSVs: one for the emotions and AUs. These files contained frame-wise data on facial expressivity and AU metrics, respectively. AU presence is a binary indicator that denotes whether a specific AU is active (1) or not (0) in a given frame. AU intensity, on the other hand, provides a more nuanced measure of emotional expression by indicating the strength of the AU activation on a continuous scale. The intensity values typically range from 0 to 1, where 0 means the AU is not present, and 1 indicates the maximum intensity of the AU.

For this study, we focused on AU intensity values as they provide a more subtle measure of emotional expression, capturing the gradations in facial movements that are crucial for identifying patterns associated with depression.

Given the structure of our experiments, we managed the timing information provided by the mobile application to trim the CSV files. This mobile app recorded the start and end times of the experiment as well as the timestamps for different experimental phases. For this study, we specifically analyzed data from the `emotional induction' (EI) phase. This phase involved inducing negative moods and emotions in participants using negative statements. By analyzing participants' expressive behavior under these standardized conditions, we gathered critical data on how their facial expressions responded to negative emotional stimuli. 

\subsubsection{Depression-Indicative Action Units}

Based on previous studies, several AUs have been identified as significantly associated with depression. These AUs were found to be more prominent or frequently activated in depressed patients compared to non-depressed patients \cite{r5} \cite{r6}.

\begin{table}[H]
\centering
\caption{Top 5 Action Units Associated with Depression}
\label{tab:top_aus_depression}
\begin{tabular}{c|c|l}
\toprule
\textbf{Rank} & \textbf{Action Unit} & \textbf{Description} \\ \midrule
1 & AU1 & Inner Brow Raiser  \\ \hline
2 & AU4 & Brow Lowerer \\ \hline
3 & AU15 & Lip Corner Depressor \\ \hline
4 & AU6 & Cheek Raiser  \\ \hline
5 & AU10 & Upper Lip Raiser \\ \bottomrule
\end{tabular}
\end{table}

According to Table \ref{tab:top_aus_depression}, the top 5 action units associated with depression include AU1 (Inner Brow Raiser), AU4 (Brow Lowerer), AU15 (Lip Corner Depressor), AU6 (Cheek Raiser), and AU10 (Upper Lip Raiser) \cite{r5, r6}. These AUs have been identified in previous research as being more frequently activated or prominent in individuals with depression compared to non-depressed individuals.

\begin{figure}[H]
\centering
\includegraphics[width=0.8\linewidth]{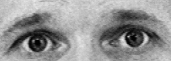}
\caption{Example of AU1: Inner Brow Raiser \cite{r7}}
\label{fig:au1_ex}
\end{figure}

\begin{figure}[H]
\centering
\includegraphics[width=0.8\linewidth]{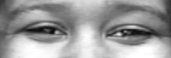}
\caption{Example of AU6: Cheek Raiser \cite{r7}}
\label{fig:au6_ex}
\end{figure}

\begin{figure}[H]
\centering
\includegraphics[width=0.8\linewidth]{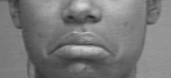}
\caption{Example of AU15: Lip Corner Depressor \cite{r7}}
\label{fig:au15_ex}
\end{figure}

\begin{table}[h]
\centering
\caption{Mapping of AUs to Emotional Expressions \cite{r8}}
\label{tab:emotion_au_mapping}
\begin{tabular}{c|l}
\toprule
\textbf{Emotion} & \textbf{Combination of AUs} \\ \midrule
Happiness & AU6 + AU12 \\ \hline
Sadness & AU1 + AU4 + AU15 \\ \hline
Surprise & AU1 + AU2 + AU5B + AU26 \\ \hline
Fear & AU1 + AU2 + AU4 + AU5 + AU20 + AU26 + AU28 \\ \hline
Disgust & AU9 + AU15 + AU16 \\ \hline
Anger & AU4 + AU5 + AU7 + AU23 \\ \bottomrule
\end{tabular}
\end{table}

Studies already mentioned in Section \ref{section:studies} have demonstrated that AUs mentioned in \ref{tab:emotion_au_mapping} are more frequently observed or have higher intensity in individuals with depression. For instance, AU1 (Inner Brow Raiser) and AU4 (Brow Lowerer) are associated with sadness and distress, while AU15 (Lip Corner Depressor) is directly linked to expressions of sadness and despair.

Using FACS, we mapped specific combinations of AUs to distinct emotional expressions. This mapping allowed us to quantify the presence and intensity of various emotions based on the AU intensity values recorded in the exported files. Table \ref{tab:emotion_au_mapping} outlines the combinations of AUs used to identify each emotion. For each frame, we calculated the intensity of these emotions by summing the intensities of the corresponding AUs. This approach provided a continuous measure of emotional expression over time. 

\subsubsection{Feature Extraction}

The primary features extracted for analysis included:

\begin{itemize}
\item AU Intensity Values: The intensity levels of individual AUs recorded for each frame of the video data.
\item Emotional Expressivity: Calculated as the cumulative intensity of specific combinations of AUs associated with distinct emotional states.
\end{itemize}

These features were pivotal in capturing both subtle and pronounced changes in facial expressions, crucial for distinguishing between depressed and healthy individuals during the emotional induction phase. 

\subsection{Mean Intensity Comparison}

To compare the differences in emotional expressions between depressed and healthy participants, we conducted a mean intensity comparison across selected emotions. This involved averaging AU intensity values over all frames during the phase for each participant.

Statistical tests were applied to compare the mean intensity values of AUs associated with emotions such as happiness and sadness, particularly between the different patient groups. Significant differences were identified to highlight distinctive emotional expression patterns. 

\subsection{PCA Clustering}

Principal Component Analysis (PCA) was employed to reduce the dimensionality of AU intensity data. This technique transformed the original set of correlated AUs into a smaller set of uncorrelated principal components (PCs). The first few PCs, which captured the highest variance in the data, were selected for subsequent clustering analysis.

Various clustering methods, including K-means clustering, Agglomerative clustering, and Gaussian Mixture Model (GMM), is applied to these principal components. These methods enabled the identification of coherent clusters of facial expression patterns, facilitating an unsupervised understanding of emotional expression variability among participants. \\

\subsection{Silhouette Score Analysis}

The quality of clusters generated by K-means clustering was assessed using the silhouette score \cite{sil_score}, a metric that evaluates the coherence and separation of clusters. For each data point, the silhouette score computed how similar it was to its cluster compared to neighboring clusters.

For each data point, the silhouette score was calculated as follows:
\[
s(i) = \frac{b(i) - a(i)}{\max(a(i), b(i))}
\]
where \(a(i)\) is the average distance from the data point \(i\) to the other points in the same cluster, and \(b(i)\) is the average distance from \(i\) to points in the nearest different cluster. 

The overall silhouette score was calculated as the mean of individual scores across all data points. Higher silhouette scores indicated well-defined clusters, providing validation of the clustering approach and insight into the distinctiveness of facial expression patterns associated with depression.

\subsection{Time Series Classification Performance}

Given the temporal nature of facial expression data, we explored the use of time series classification to distinguish between depressed and healthy patients. Each video sequence was treated as a time series of AU intensity values. 

A crucial step before applying any classification model is data pre-processing. This typically involves splitting the data into training (80\% of the whole dataset) and testing sets (20\% of the whole dataset). Splitting strategies include random splitting and stratified splitting (maintaining class proportions in both sets). Scaling the data to a common range (e.g., 0-1 or standard deviation) to ensure all features contribute equally during model training. \\
    We employed several classification models tailored for time series analysis:

\begin{itemize}
    \item ROCKET + LogisticRegression: This method involves applying random convolutional kernels to the time series to extract features efficiently. After feature extraction, a logistic regression classifier is trained on the transformed features \cite{rocket}.
    \item ROCKET + RidgeClassifierCV: Similar to Logistic Regression but uses ridge regression for classification, which can handle multicollinearity among features better \cite{rocket}.
    \item InceptionTime: A deep learning architecture designed specifically for time series classification, leveraging inception modules similar to those used in image classification \cite{inception}.
    \item LSTM: A type of recurrent neural network (RNN) well-suited for sequence data, capable of learning long-term dependencies in time series \cite{lstm}.
    \item XGBoost Classifier: An ensemble learning method known for its efficiency and effectiveness in various machine learning tasks, including time series classification \cite{xgb}.
\end{itemize}

Performance was evaluated using accuracy to assess the model's ability to classify sequences correctly. 

\section{Results and Discussion}

\subsection{Descriptive Statistics}

\begin{figure}[H]
\centering
\includegraphics[width=0.8\linewidth]{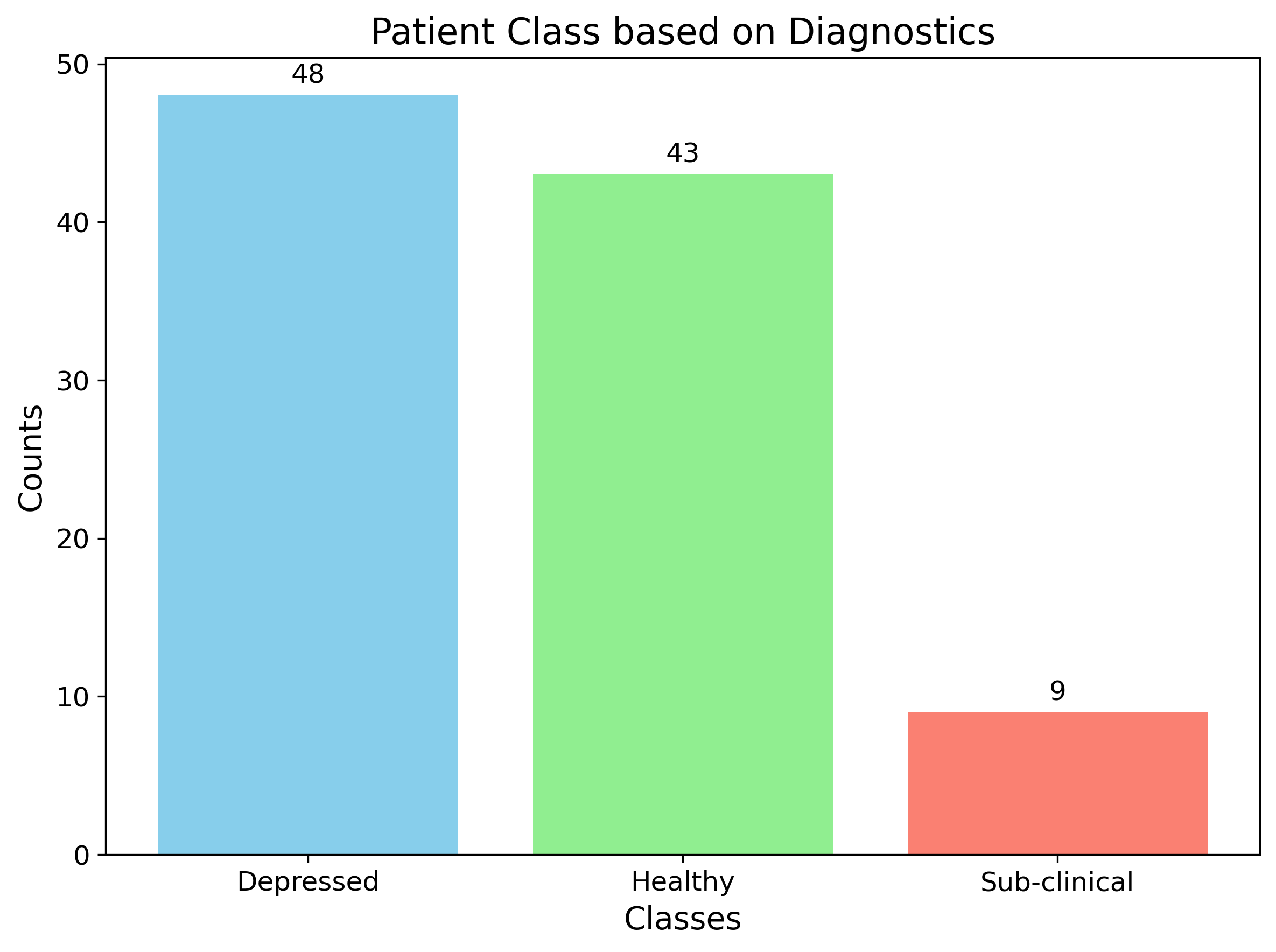}
\caption{Distribution of participants across three categories in the EmpkinS D02 dataset: Depressed (\(n=46\)), Healthy (\(n=43\)), and Sub-clinical (\(n=9\)).}
\label{fig:patient_class}
\end{figure}

The bar plot in Figure \ref{fig:patient_class} shows the distribution of participants in our study across three categories: Depressed, Healthy, and Sub-clinical. A smaller group of participants (9 individuals) is categorized as Sub-clinical, indicating that they exhibited some signs of depression but did not meet the full criteria for a depressed classification.

To ensure a more comprehensive analysis, these sub-clinical participants were included in the Depressed category for subsequent analysis. This decision was made based on their demonstrated symptoms of depression, which justifies their inclusion in the depressed group to better understand the spectrum of depressive symptoms and their impact on facial expressivity and emotional expression metrics. This reclassification resulted in a combined Depressed group of 57 participants, providing a more robust dataset for our analysis. 

\subsection{Mean Intensity Comparison of the Dominant Action Units}

In this series of experiments, we compared the mean intensity of the dominant AUs associated with depression between depressed and healthy patients which are known to be significantly associated with facial expressions related to depression, offering a measurable basis for comparing facial expressivity between the two groups, specifically, Inner Eyebrow Raise (AU1), Brow Lowerer (AU4), and Lip Corner Depressor (AU15). The comparison of these AUs provides insight into the facial expressivity differences between the two groups over time during the target phase.

\begin{figure}[H]
\centering
\includegraphics[width=0.8\linewidth]{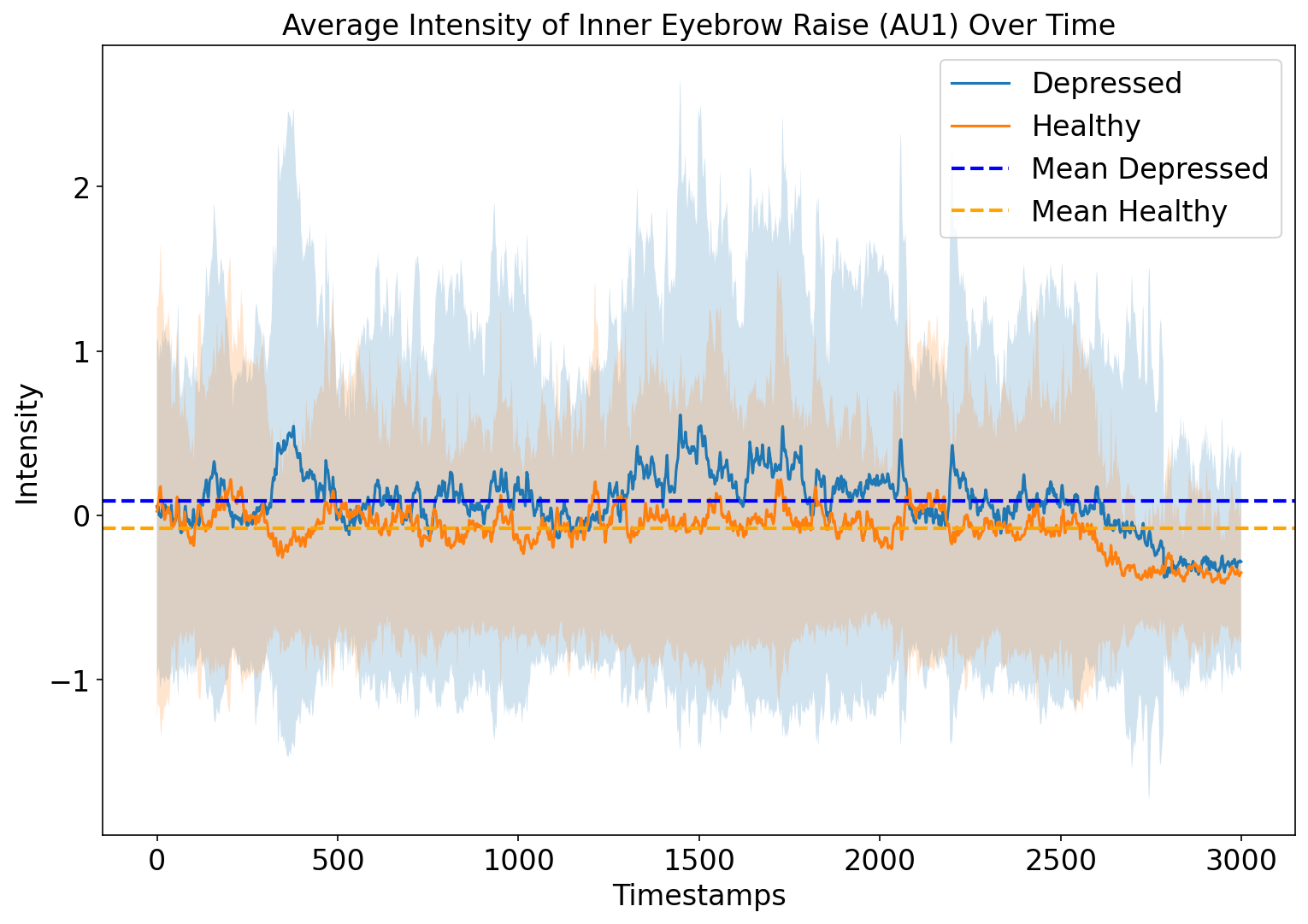}
\caption{Comparison of Average Intensity of Inner Eyebrow Raise (AU1) Over Time Between Depressed and Non-Depressed Patients.}
\label{fig:au1}
\end{figure}

\begin{figure}[H]
\centering
\includegraphics[width=0.8\linewidth]{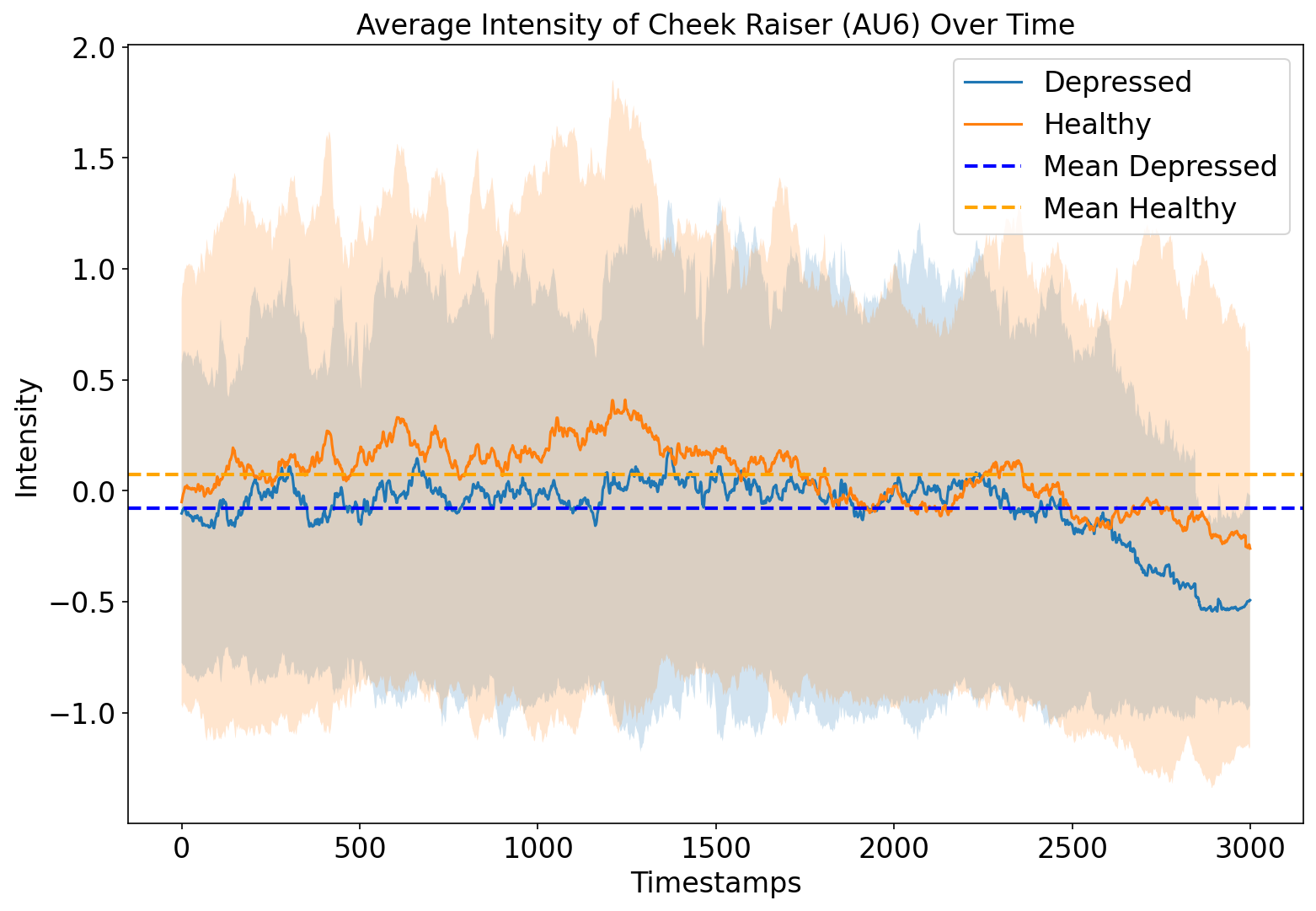}
\caption{Comparison of Average Intensity of Cheek Raiser (AU6) Over Time Between Depressed and Non-Depressed Patients.}
\label{fig:au6}
\end{figure}

\begin{figure}[H]
\centering
\includegraphics[width=0.8\linewidth]{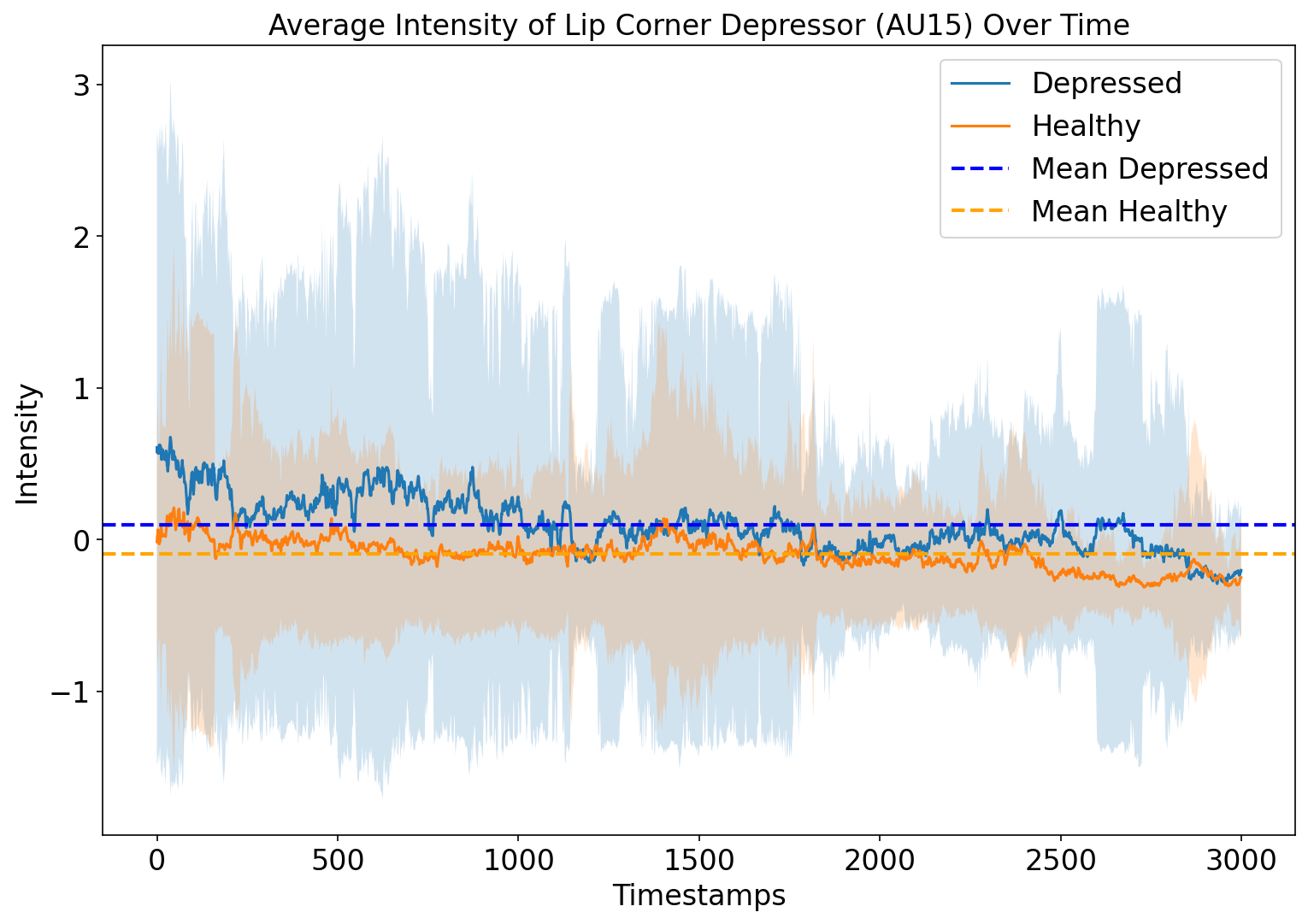}
\caption{Comparison of Average Intensity of Lip Corner Depressor (AU15) Over Time Between Depressed and Non-Depressed Patients.}
\label{fig:au15}
\end{figure}

The plots \ref{fig:au1}, \ref{fig:au6} \& \ref{fig:au15} include a horizontal dashed line representing the overall mean intensity of happiness for both groups, as well as shaded areas depicting the standard deviation. These additions provide a clear visual representation of the variability and central tendency of the data.

As illustrated in Figures \ref{fig:au1}, \ref{fig:au6}, and \ref{fig:au15}, there are notable differences in the mean intensity of specific AUs between depressed and healthy patients.

\begin{itemize}
    \item Inner Eyebrow Raise (AU1): Figure \ref{fig:au1} shows that the average intensity of AU1 is generally higher in depressed patients compared to healthy patients. This difference is consistent over time, suggesting that depressed patients exhibit more pronounced inner eyebrow raises, often associated with sadness and distress.
    \item Cheek Raiser (AU6): Figure \ref{fig:au6} demonstrates that the average intensity of AU6 is generally higher in depressed patients compared to healthy patients. This difference is noticeable over time, indicating that depressed patients exhibit more pronounced brow lowering, which is associated with expressions of sadness and concern.
    \item Lip Corner Depressor (AU15): As shown in Figure \ref{fig:au15}, the average intensity of AU15 is significantly higher in depressed patients. This action unit is associated with expressions of sadness and despair, aligning with the emotional state of depressed individuals.
\end{itemize}

\begin{table}[!htbp]
\centering
\begin{tabular}{l|c|c|c}
\toprule
 \textbf{Action Units} & \textbf{AU1} & \textbf{AU6} & \textbf{AU15} \\
\midrule
\textbf{Depressed} & 0.087 & -0.077 & 0.100 \\ \midrule
\textbf{Healthy} & -0.081 & 0.071 & -0.092 \\  \midrule
\textbf{Difference} & 0.167 & -0.148 & 0.192 \\ \bottomrule
\end{tabular}
\caption{Overall Mean Intensity of Dominant AUs in Depressed and Healthy Patients.}
\label{tab:mean_intensity}
\end{table}

The analysis of mean intensity comparisons of key action units (AU1, AU4, AU15) in Table \ref{tab:mean_intensity} reveals substantial differences in facial expressivity between depressed and healthy patients. Depressed individuals consistently exhibit higher intensities of these action units, corresponding to expressions of sadness, distress, and concern. These findings support the hypothesis that specific facial expressions can serve as biomarkers for depression, providing a potential avenue for more objective and automated psychological assessments. 

\subsection{Mean Intensity Comparison of Dominant Expressions: Sadness and Happiness}

In this section, we analyze the mean intensity of dominant expressions, specifically sadness and happiness, between depressed and healthy patients. This comparison highlights the differences in emotional expressivity between the two groups, providing further insight into their emotional states.

\begin{figure}[!htbp]
\centering
\includegraphics[width=0.8\linewidth]{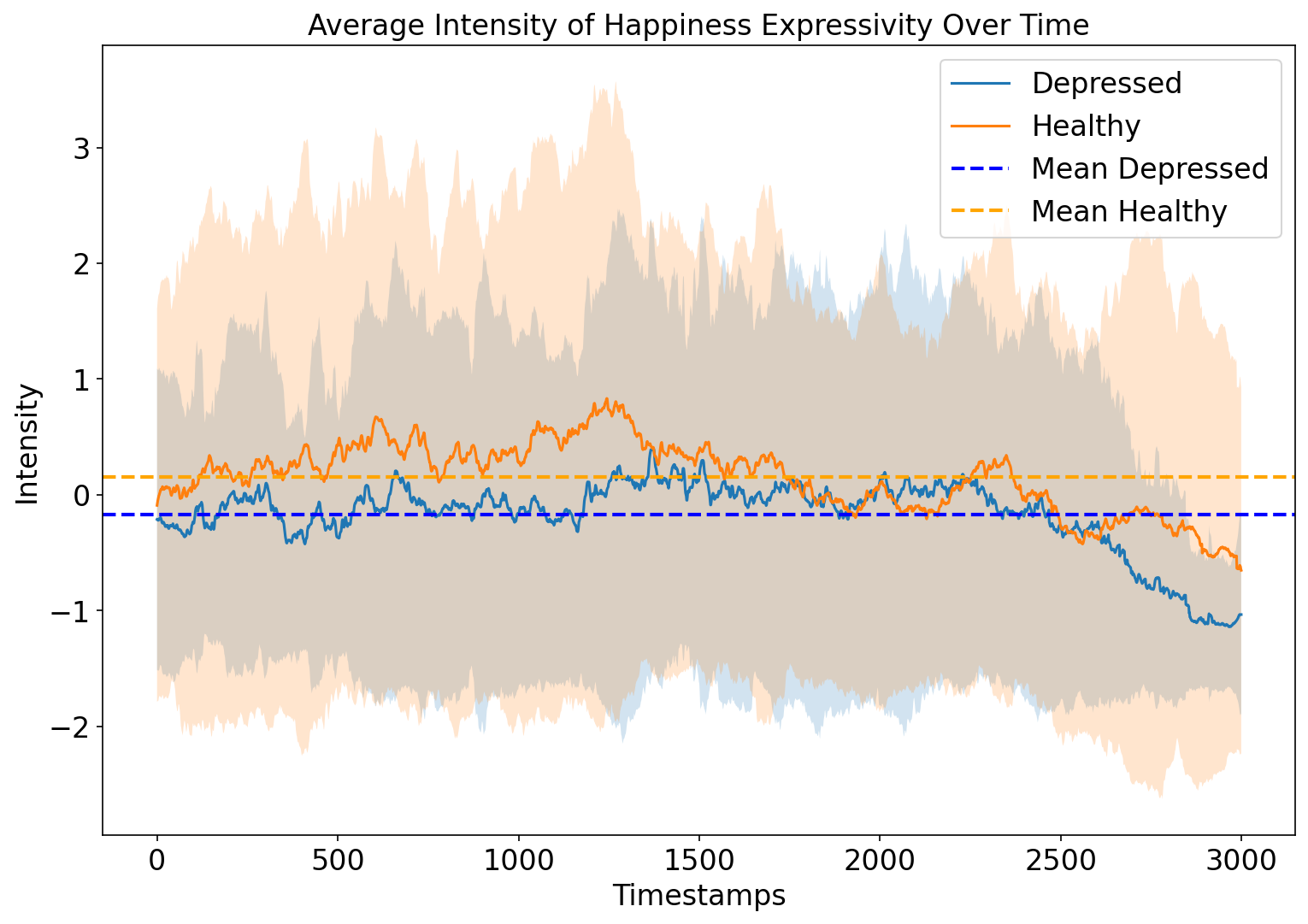}
\caption{Comparison of Average Intensity of Happiness Over Time Between Depressed and Healthy Patients.}
\label{fig_hap}
\end{figure}

\begin{figure}[!htbp]
\centering
\includegraphics[width=0.8\linewidth]{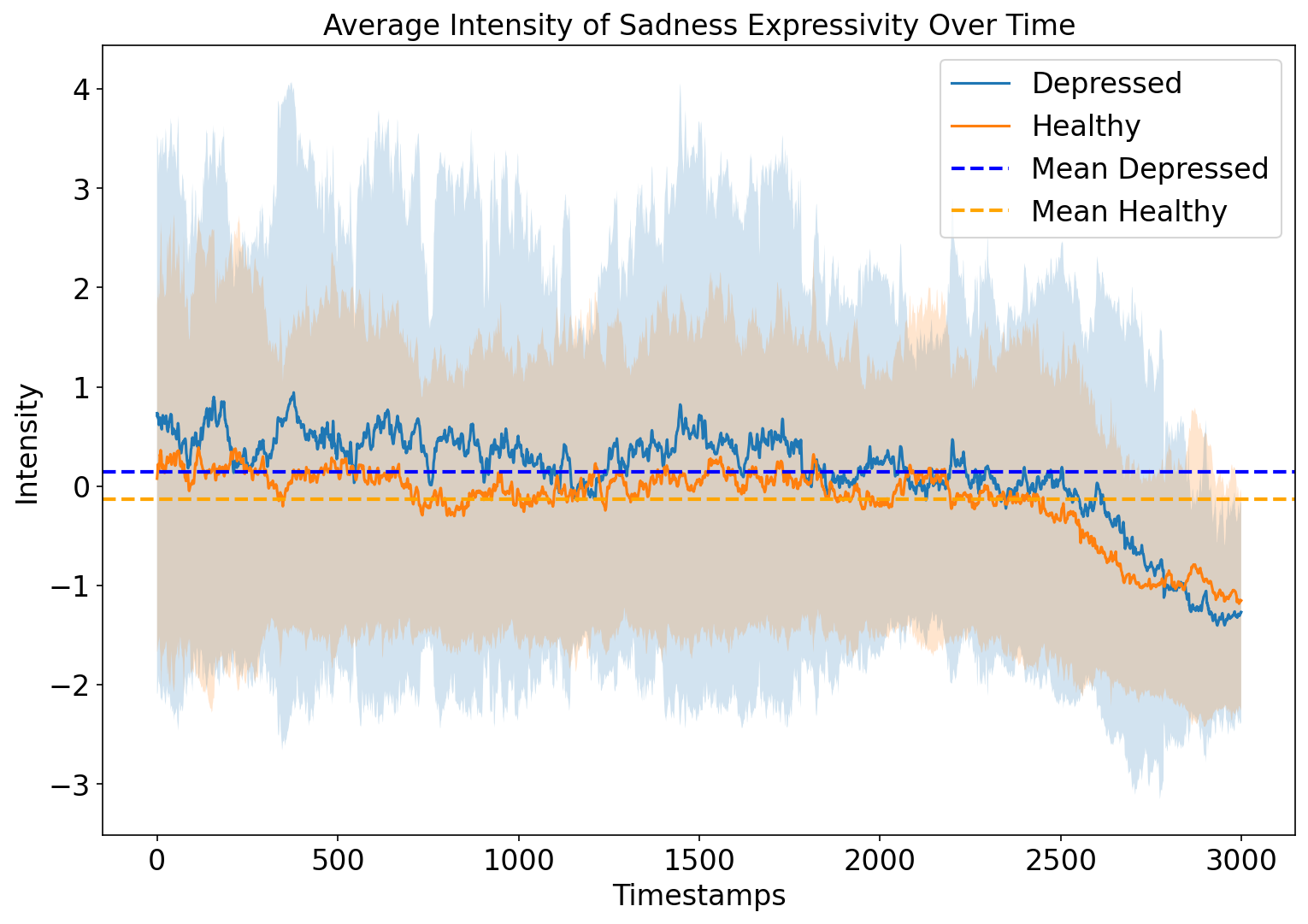}
\caption{Comparison of Average Intensity of Sadness Over Time Between Depressed and Healthy Patients.}
\label{fig_sad}
\end{figure}

Figures \ref{fig_hap} and \ref{fig_sad} illustrate the mean intensity of expressions of happiness and sadness, respectively, over time.

\begin{enumerate}
\item \textbf{Happiness}: As shown in Figure \ref{fig_hap}, the average intensity of happiness is consistently lower in depressed patients compared to healthy patients. This suggests that depressed individuals exhibit fewer and less intense expressions of happiness, indicative of their reduced positive affect.
\item \textbf{Sadness}: Figure \ref{fig_sad} shows that the average intensity of sadness is markedly higher in depressed patients compared to healthy patients. This consistent difference underscores the prevalence of negative affect and emotional distress in depressed individuals.
\end{enumerate}

The analysis of dominant expressions reveals that depressed patients show significantly lower intensities of happiness and higher intensities of sadness. These findings align with the expected emotional profiles of depressed individuals, reinforcing the potential of using facial expressions as objective indicators for assessing depression. 

\subsection{PCA and Clustering Analysis}

\subsubsection{Explained Variance by Principal Components}

\begin{figure}[!htbp]
\centering
\includegraphics[width=0.8\linewidth]{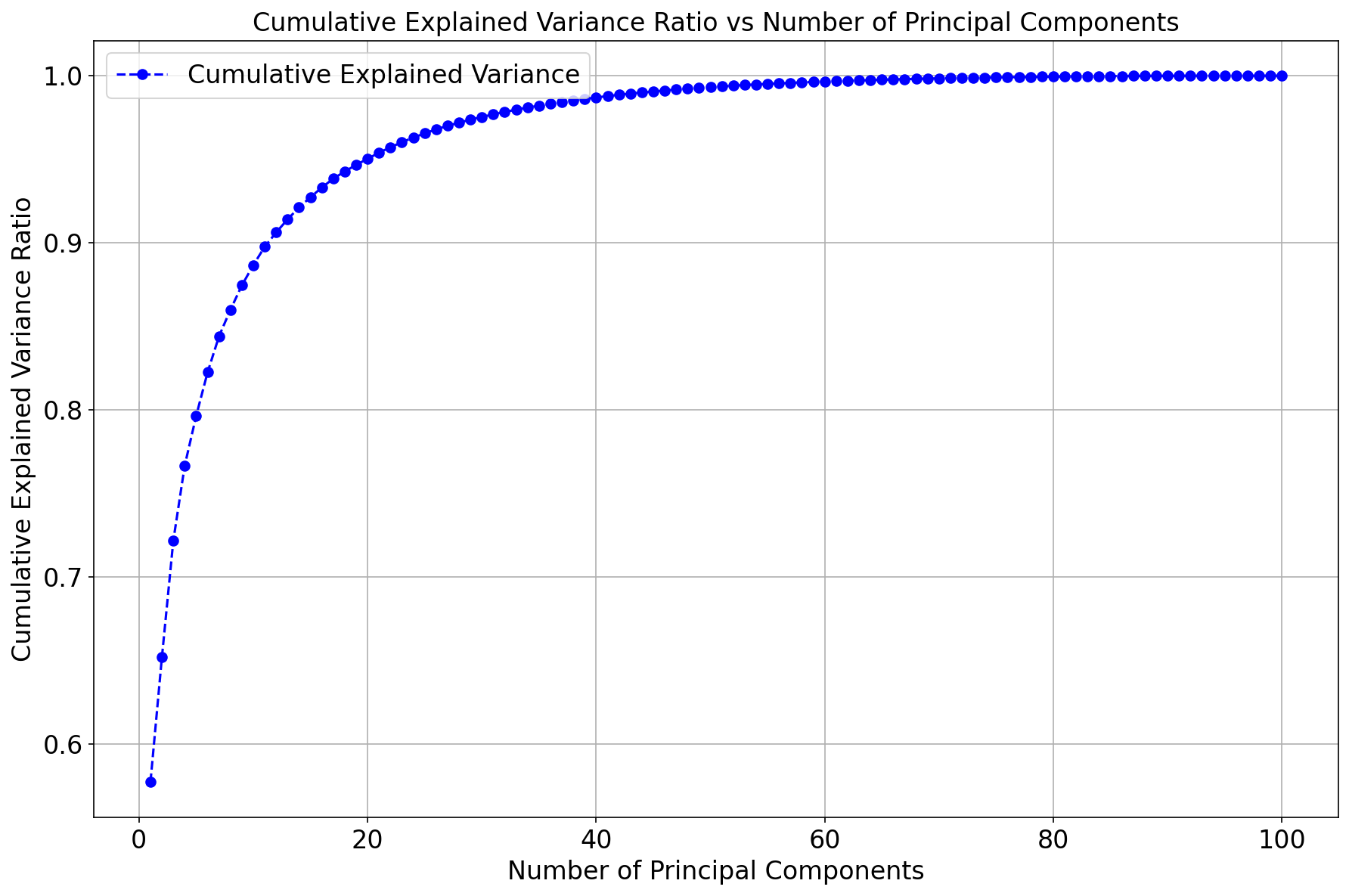}
\caption{Cumulative variance explained by PCA with 20 components selected to retain 95\% of variance.}
\label{fig:pca_cumulative}
\end{figure}

Figure \ref{fig:pca_cumulative} illustrates the cumulative explained variance with respect to the number of principal components, demonstrating that the first 20 components explain approximately 95\% of the variance.

\subsubsection{Clustering Analysis}

In this section, we perform K-means clustering on PCA-transformed data to analyze the clustering patterns of emotion intensities. We applied K-Means clustering with 2 clusters on the PCA-transformed data. The silhouette score was computed to evaluate the clustering performance and is noted in Section \ref{section:sil_scores}.

\begin{figure}[!htbp]
\centering
\includegraphics[width=0.8\linewidth]{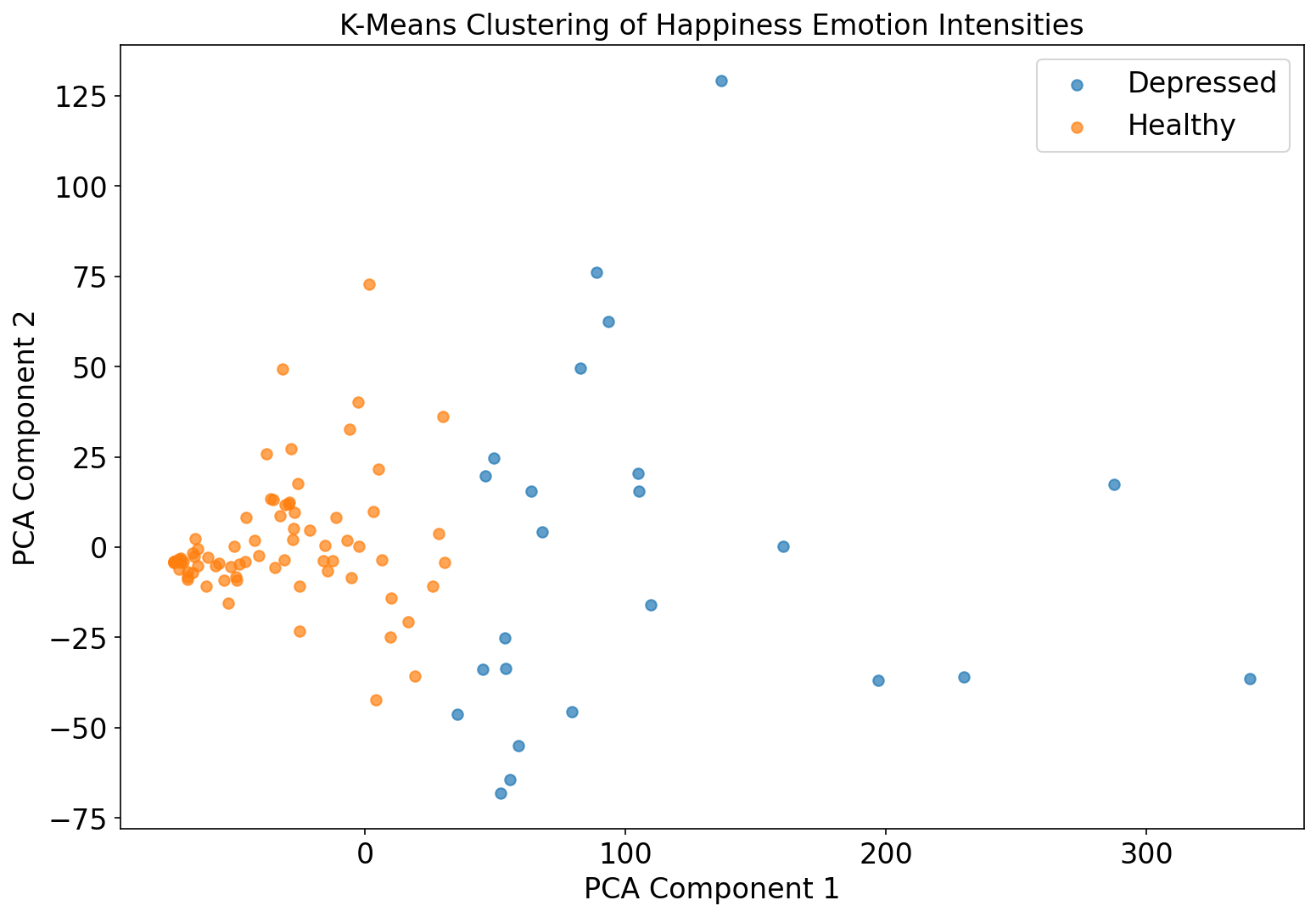}
\caption{K-Means Clustering of Happiness Emotion Intensities.}
\label{fig:hap_kmeans}
\end{figure}

\begin{figure}[!htbp]
\centering
\includegraphics[width=0.8\linewidth]{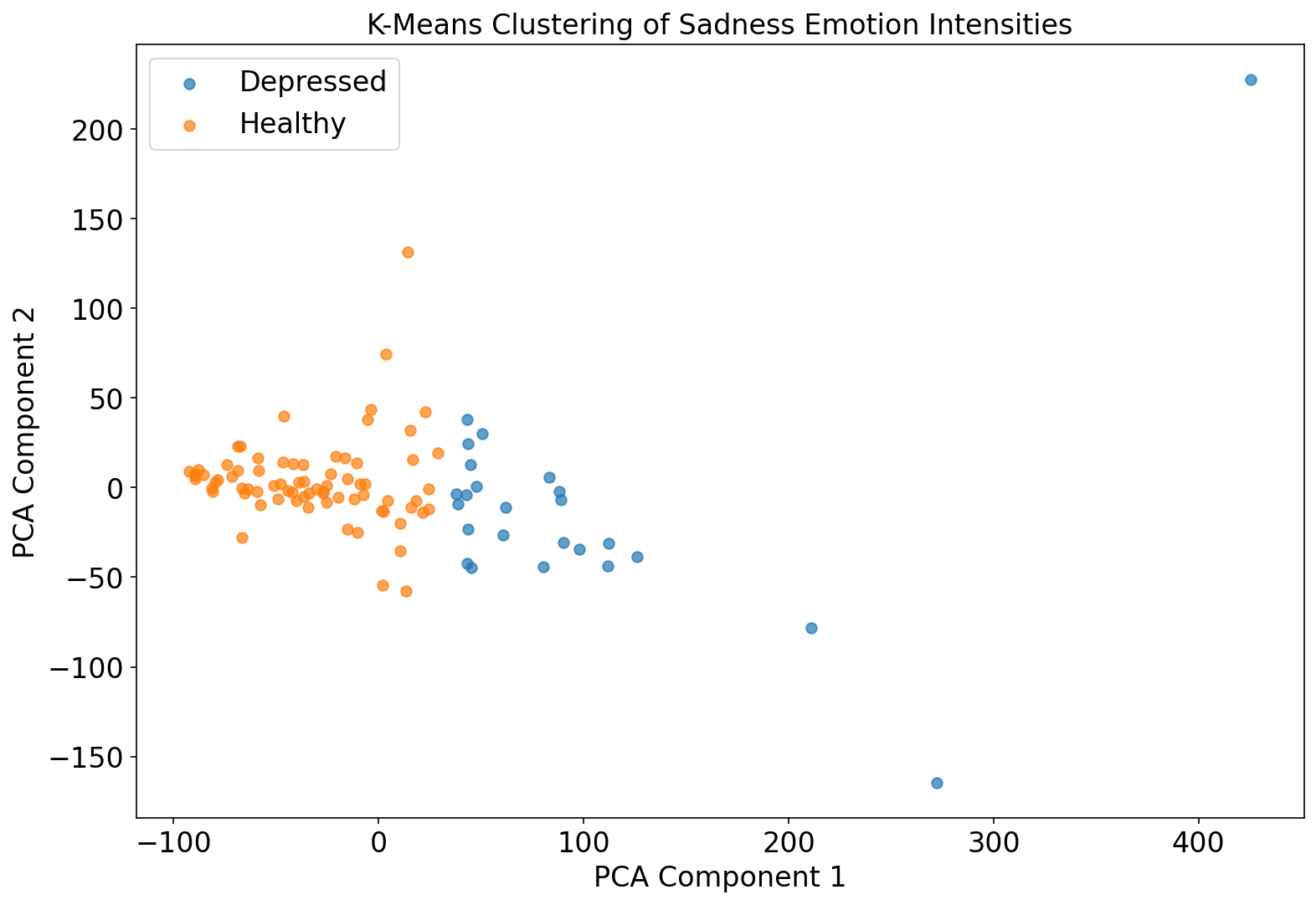}
\caption{K-Means Clustering of Sadness Emotion Intensities.}
\label{fig:sad_kmeans}
\end{figure}

In Figure \ref{fig:hap_kmeans} \& \ref{fig:sad_kmeans} we see the results of K-means clustering applied to PCA-transformed data. It visualizes how data points are grouped into clusters based on the similarity of emotion intensity features. The clusters are reasonably well-separated and most data points are correctly assigned to their clusters, with some overlap near cluster boundaries.

We then perform Agglomerative Clustering on PCA-transformed data to analyze the clustering patterns of emotion intensities. It works from the dissimilarities between the objects to be grouped.

\begin{figure}[!htbp]
\centering
\includegraphics[width=0.8\linewidth]{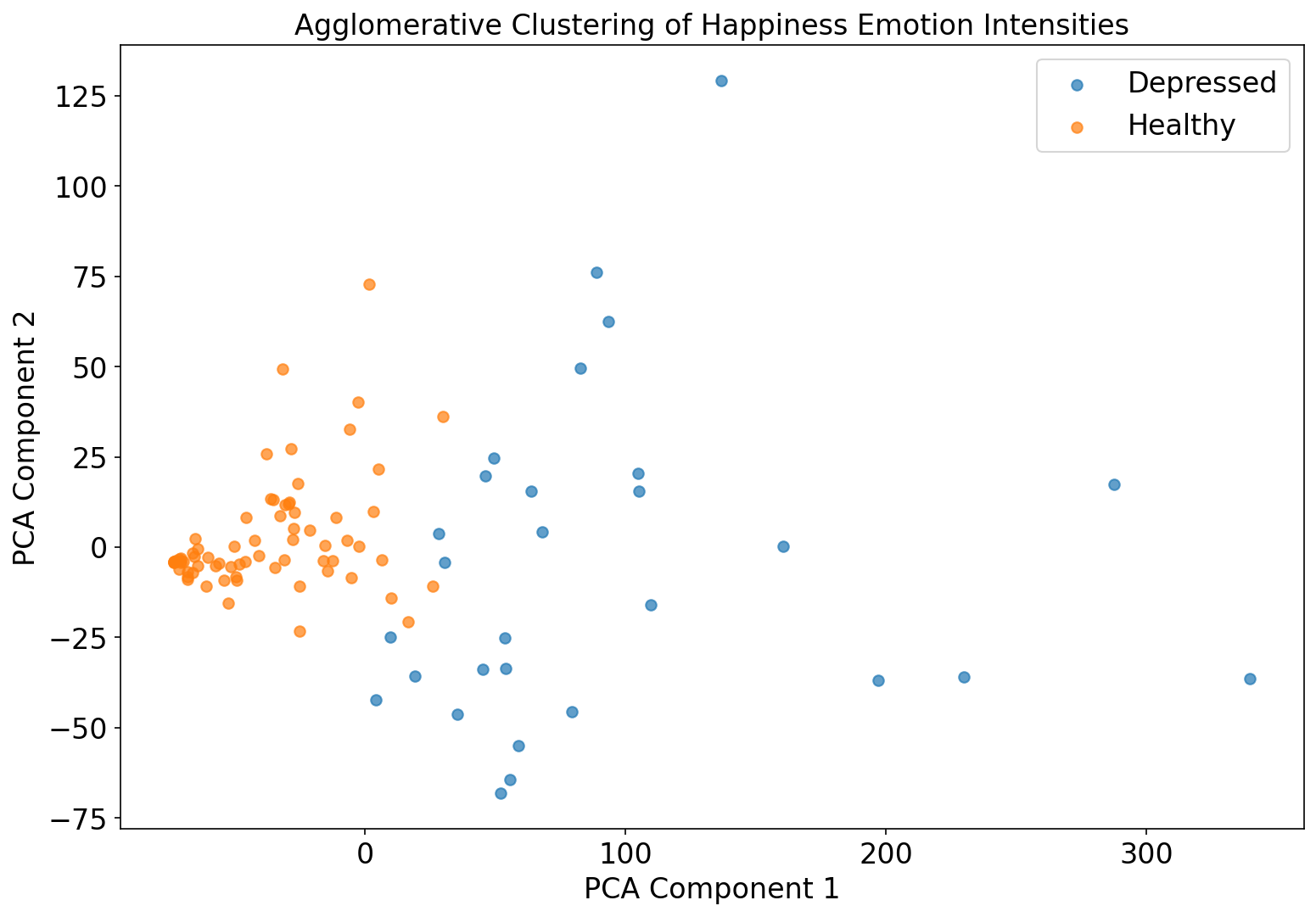}
\caption{Agglomerative Clustering of Happiness Emotion Intensities.}
\label{fig:hap_algo}
\end{figure}

\begin{figure}[!htbp]
\centering
\includegraphics[width=0.8\linewidth]{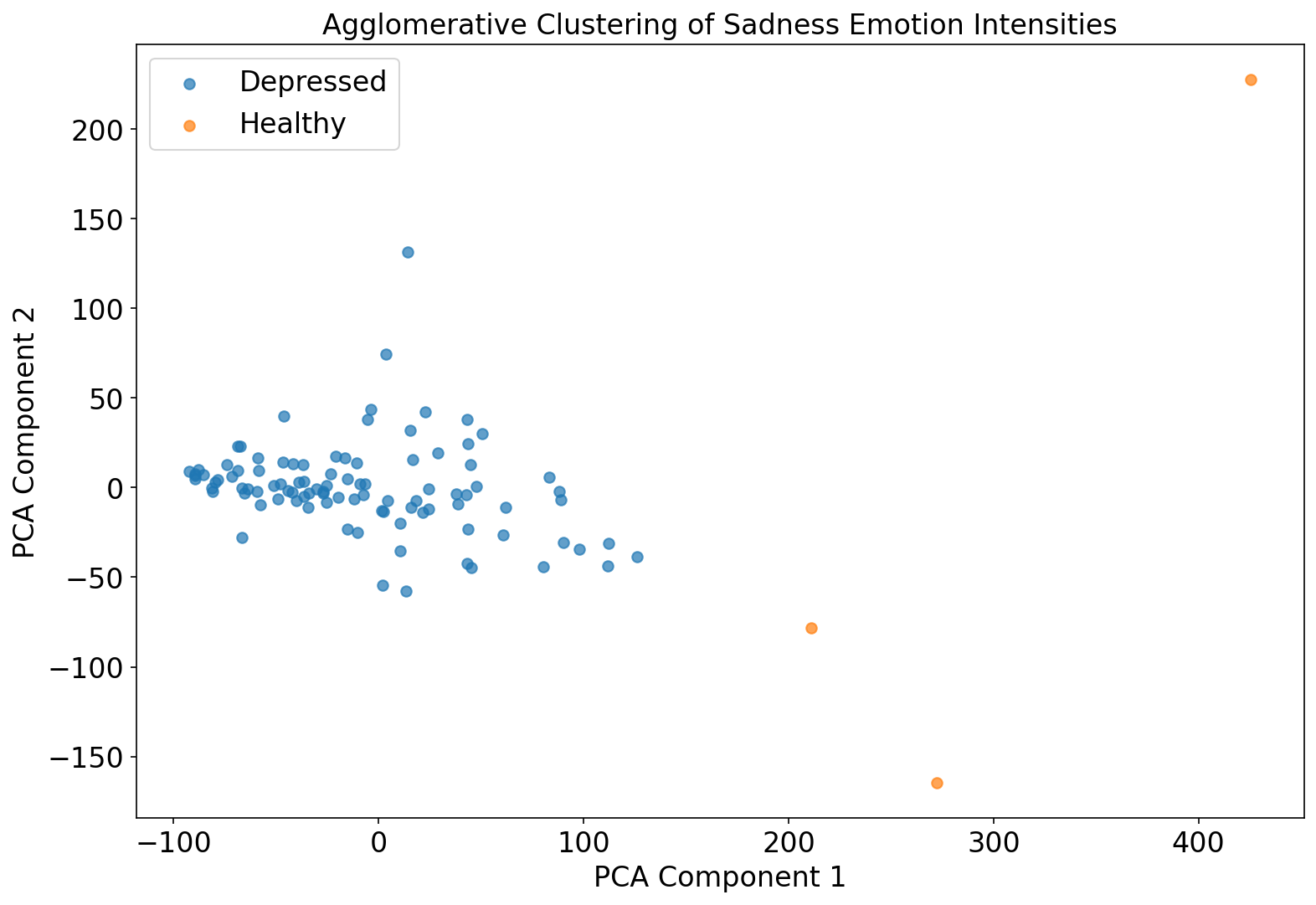}
\caption{Agglomerative Clustering of Sadness Emotion Intensities.}
\label{fig:sad_algo}
\end{figure}

In Figure \ref{fig:hap_algo} \& \ref{fig:sad_algo}, we see the results of Agglomerative clustering applied to the transformed data. The clusters are well-separated from each other, data points in one cluster are distinctly different from those in neighboring cluster where the cluster for healthy patients indeed contains significantly fewer points than the other.

We utilize the Gaussian Mixture Model (GMM) on PCA-transformed data to analyze the clustering patterns of emotion intensities.

\begin{figure}[!htbp]
\centering
\includegraphics[width=0.8\linewidth]{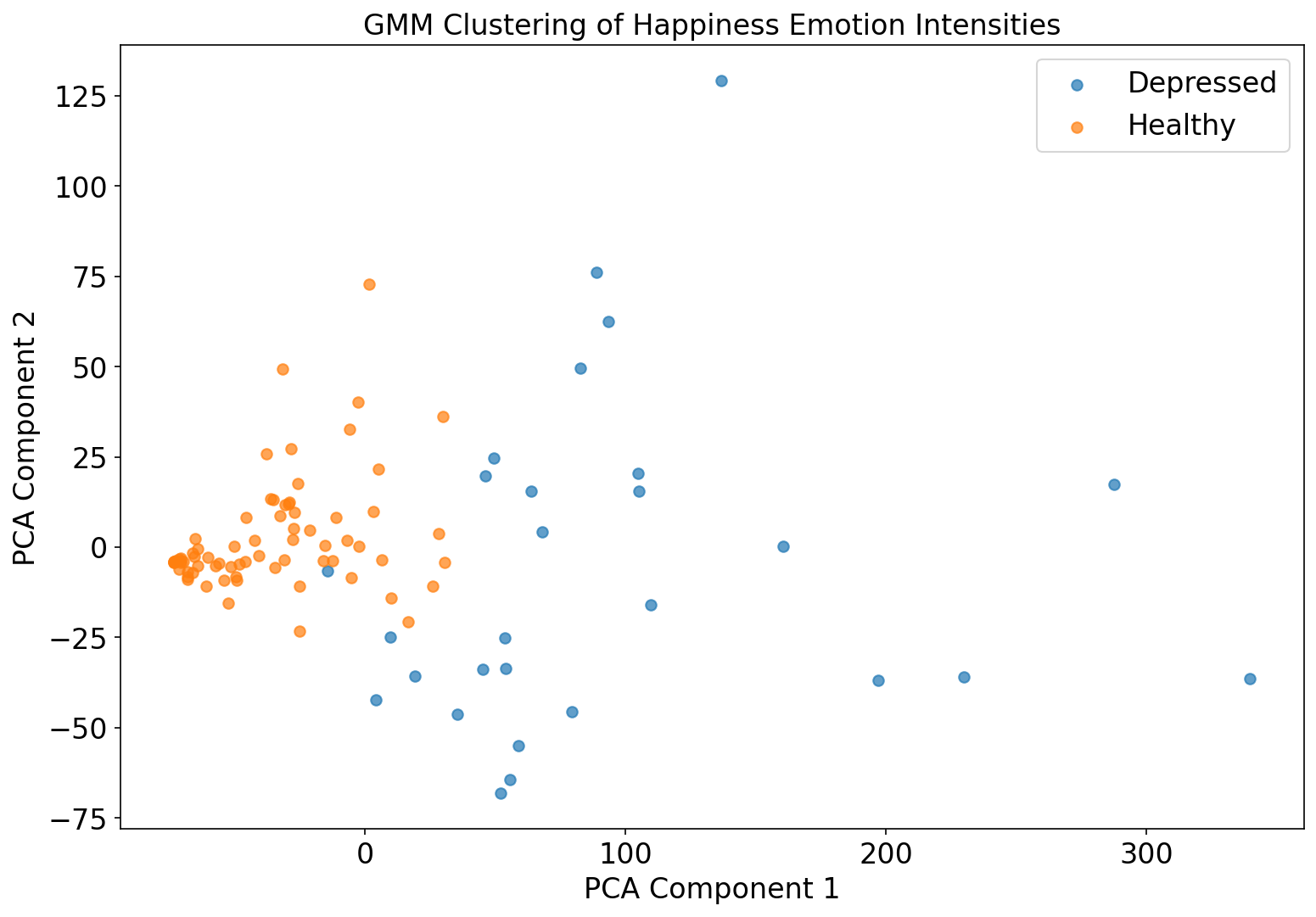}
\caption{GMM Clustering of Happiness Emotion Intensities.}
\label{fig:hap_gmm}
\end{figure}

\begin{figure}[!htbp]
\centering
\includegraphics[width=0.8\linewidth]{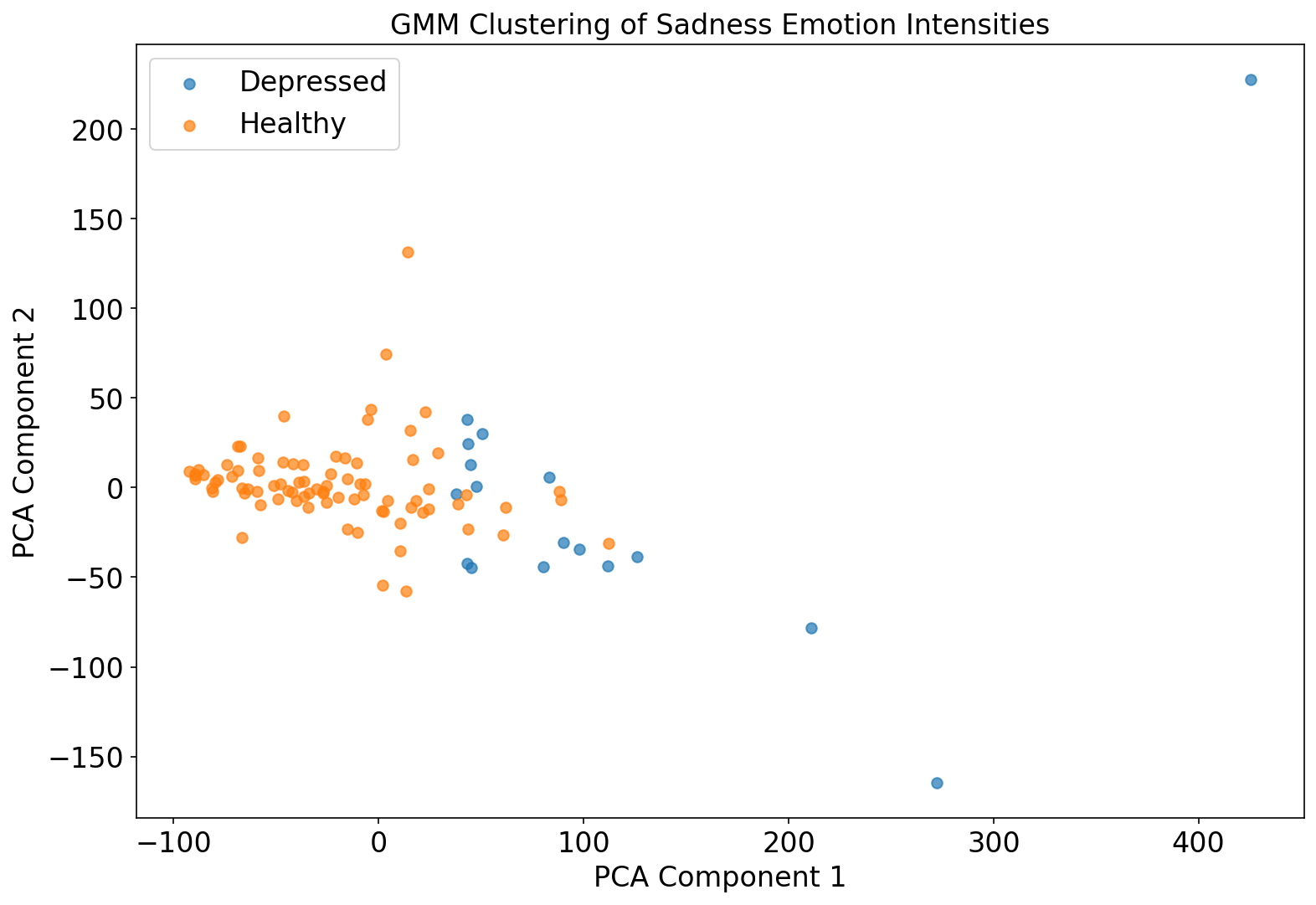}
\caption{GMM Clustering of Sadness Emotion Intensities.}
\label{fig:sad_gmm}
\end{figure}

In Figure \ref{fig:hap_gmm} \& \ref{fig:sad_gmm} we see the results of Agglomerative clustering applied to the transformed data. The clusters are distinguishable from each other, but there is some overlap or ambiguity in the assignment of a few data points.

\subsubsection{Evaluating Silhouette Scores} \label{section:sil_scores}

Table \ref{tab:silhouette_scores} compares the silhouette scores for happiness and sadness across different clustering algorithms.

\begin{table}[!htbp]
\centering
\begin{tabular}{c|c|c}
\toprule
\textbf{Clustering Algorithm} & \textbf{Score (Happiness)} & \textbf{Score (Sadness)} \\
\midrule
K-means & 0.478 & 0.417 \\
\hline
Agglomerative & 0.446 & 0.709 \\
\hline
GMM & 0.437 & 0.431 \\
\bottomrule
\end{tabular}
\caption{Comparison of Silhouette Scores for Different Clustering Algorithms}
\label{tab:silhouette_scores}
\end{table}

Based on the comparison of silhouette scores for happiness and sadness intensities across different clustering algorithms (Table \ref{tab:silhouette_scores}), Agglomerative Clustering emerges as the optimal method, achieving the highest score of 0.709 and 0.446. It demonstrates a moderately well-defined clustering structure. A score in this range suggests that the clusters are sufficiently distinct and separated from each other, although there may still be some overlap or ambiguity at the boundaries between clusters. These results underline the effectiveness of Agglomerative Clustering in capturing distinct emotional patterns, particularly for sadness, thereby showcasing its suitability for such clustering tasks.

\subsection{Time Series Classification Performance}

This section evaluates the performance of various classification techniques on our dataset, which consists of emotion intensities related to happiness and sadness. We compare the accuracy scores achieved by each method and discuss key aspects of their implementation.

We utilized the learning rate finder from the 'tsai' framework \cite{tsai}, a leading deep learning library for time series and sequences, to ascertain the optimal learning rate for training data across all classification techniques employed.

\begin{figure}[!htbp]
\centering
\includegraphics[width=0.7\linewidth]{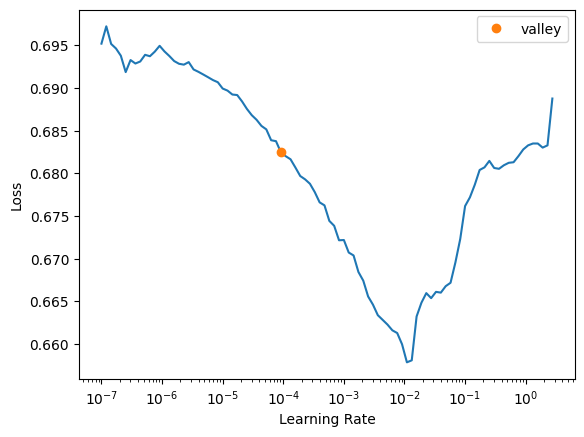}
\caption{Example - Learning Rate Finder for InceptionTime.}
\label{fig:lr_finder}
\end{figure}

A comparison of accuracy scores achieved by each method on the dataset is presented in Table \ref{tab:accuracy_comparison}.\\

\begin{table}[!htbp]
\centering
\begin{tabular}{@{}lc@{}}
\toprule
\textbf{Method}                     & \textbf{Accuracy Score} \\ \midrule
ROCKET + Logistic Regression        & 0.70                    \\
ROCKET + RidgeClassifierCV          & 0.66                    \\
InceptionTime                       & 0.60                    \\
LSTM                                & 0.51                    \\
XGBoost Classifier                  & 0.54                    \\ \bottomrule
\end{tabular}
\caption{Comparison of accuracy scores using various time series classification techniques.}
\label{tab:accuracy_comparison}
\end{table}

\begin{figure}[!htbp]
\centering
\includegraphics[width=0.8\linewidth]{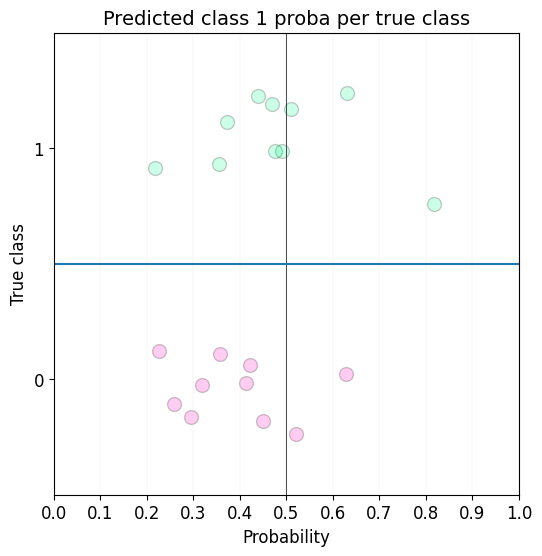}
\caption{Plot for predicted class probability per true class for ROCKET + LogisticRegression.}
\label{fig:proba}
\end{figure}

Based on the results in Table \ref{tab:accuracy_comparison}, ROCKET (Random Convolutional Kernel Transform) combined with Logistic Regression demonstrates superior performance in our classification task potentially due to:

\begin{itemize}
    \item Effective Feature Extraction: Utilizes random convolutional kernels to capture diverse time-dependent patterns, providing crucial features for accurate classification.
    
    \item Dimensionality Reduction: Reduces high-dimensional feature space through techniques like random projections and pooling, preserving essential information and enhancing generalization.
    
    \item Efficient Training: Logistic Regression efficiently learns linear decision boundaries in the transformed feature space, enabling faster training compared to complex models. \\
\end{itemize} 

\section{Conclusion}

This study demonstrates the efficacy of using facial action units (AUs) and emotions as objective biomarkers for detecting depression. The analysis of facial action units, specifically comparing their mean intensities, supports the hypothesis that these metrics effectively evaluate a patient's condition. The results also highlight sadness and happiness as the predominant emotions observed during patient evaluations.

By applying time series classification to facial expression data, significant differences in the intensity of specific AUs between depressed and healthy individuals were identified. Moving forward, future research should concentrate on refining these models and exploring multi-modal approaches that integrate facial expression analysis with other behavioral data sources such as voice and text. This integration aims to enhance diagnostic accuracy and reliability, offering a more comprehensive understanding of an individual's mental health status and enabling personalized and timely interventions.

The findings suggest that automated facial analysis can complement traditional diagnostic methods, providing a more objective and non-invasive approach to mental health assessment. Future studies should continue to refine these models and explore multi-modal frameworks to further improve diagnostic efficacy and reliability in clinical settings. \\

\section*{Acknowledgments}
This work was funded by the Deutsche Forschungsgemeinschaft (DFG, German Research Foundation)—SFB 1483—Project-ID 442419336, EmpkinS.


\bibliographystyle{IEEEtran}
\bibliography{references}

@misc{who_report2,
author = {World Health Organization},
title = {Depression},
howpublished = {Online},
year = {2021},
note = {Available: https://www.who.int/news-room/fact-sheets/detail/depression. Accessed: 25-Jun-2024}
}

@article{dep_rating,
author = {Hamilton, M.},
title = {A rating scale for depression},
journal = {Journal of Neurology, Neurosurgery, and Psychiatry},
year = {1960},
volume = {23},
number = {1},
pages = {56--62}
}

@article{phq,
author = {Kroenke, K. and Spitzer, R. L. and Williams, J. B. W.},
title = {The PHQ-9: Validity of a Brief Depression Severity Measure},
journal = {Journal of General Internal Medicine},
year = {2001},
volume = {16},
number = {9},
pages = {606--613}
}

@article{jones2018,
author = {Jones, A. and Ngo, H. Q. and Miller, R. L.},
title = {Facial Action Units and Depression: Evidence from a Large Clinical Sample},
journal = {Journal of Affective Disorders},
year = {2018},
volume = {245},
pages = {65--72}
}

@book{facs,
author = {Ekman, P. and Friesen, W. V.},
title = {Facial Action Coding System: A Technique for the Measurement of Facial Movement},
publisher = {Consulting Psychologists Press},
address = {Palo Alto},
year = {1978}
}

@book{sadness_happiness,
author = {Ekman, F. and Rosenberg, E.},
title = {What the Face Reveals: Basic and Applied Studies of Spontaneous Expression Using the Facial Action Coding System (FACS)},
edition = {2nd},
publisher = {Oxford University Press},
year = {2005}
}

@inproceedings{facs_depression,
author = {Girard, J. M. and Cohn, J. F. and Mahoor, M. H. and Mavadati, S. M. and Rosenwald, D. P.},
title = {Social Risk and Depression: Evidence from Manual and Automatic Facial Action Unit Analysis},
booktitle = {Proceedings of the 10th IEEE International Conference and Workshops on Automatic Face and Gesture Recognition (FG)},
year = {2013},
pages = {1--7}
}

@book{beck,
author = {Beck, A. T. and Steer, R. A. and Brown, G. K.},
title = {Beck Depression Inventory–II},
publisher = {Psychological Corporation},
address = {San Antonio, TX},
year = {1996}
}

@article{r1,
author = {Li, M. and Yang, Y. and Shi, W. and Wang, B.},
title = {Classification of depression based on facial expressions using machine learning},
journal = {Journal of Affective Disorders},
year = {2020},
volume = {276},
pages = {263--271},
doi = {10.1016/j.jad.2020.07.012}
}

@article{r2,
author = {Zhang, Y. and Liu, X. and Wang, Z. and Li, H.},
title = {Hybrid model combining CNNs and RNNs for depression detection through temporal analysis of facial expressions},
journal = {IEEE Transactions on Affective Computing},
year = {2022},
volume = {13},
number = {2},
pages = {325--335},
doi = {10.1109/TAFFC.2022.3141123}
}

@article{r3,
author = {Wang, J. and Chen, L. and Zhang, X. and Yang, S.},
title = {Multi-modal deep learning for depression detection: Integrating facial expression, voice, and text analysis},
journal = {Journal of Affective Disorders},
year = {2023},
volume = {321},
pages = {246--255},
doi = {10.1016/j.jad.2023.01.112}
}

@inproceedings{r4,
author = {Baltrusaitis, T. and Zadeh, A. and Lim, Y. C. and Morency, L.-P.},
title = {OpenFace 2.0: Facial behavior analysis toolkit},
booktitle = {IEEE International Conference on Automatic Face \& Gesture Recognition (FG 2018)},
year = {2018},
pages = {59--66},
doi = {10.1109/FG.2018.00019}
}

@article{r5,
author = {Li, X. and Huang, S. and Sui, J.},
title = {Facial action units and machine learning for depression detection: A review},
journal = {IEEE Transactions on Affective Computing},
year = {2020},
volume = {11},
number = {3},
pages = {432--444},
doi = {10.1109/TAFFC.2020.2979262}
}

@article{r6,
author = {Zhang, W. and Lin, H. and Liu, Z. and Yu, J.},
title = {Hybrid deep learning models for capturing spatiotemporal features of facial expressions in depression detection},
journal = {Journal of Affective Disorders},
year = {2022},
volume = {295},
pages = {897--904},
doi = {10.1016/j.jad.2021.09.089}
}

@inproceedings{r7,
author = {Kanade, T. and Cohn, J. F. and Tian, Y.},
title = {Comprehensive database for facial expression analysis},
booktitle = {Proceedings of FG00},
year = {2000},
pages = {46--53}
}

@incollection{r8,
author = {Cohn, J. F. and Ekman, P.},
title = {Measuring facial action by manual coding, facial EMG, and automatic facial image analysis},
booktitle = {The New Handbook of Methods in Nonverbal Behavior Research},
editor = {Harrigan, J. A. and Rosenthal, R. and Scherer, K. R.},
publisher = {Oxford University Press},
year = {2005},
pages = {9--64}
}

@misc{tsai,
author = {Ignacio Oguiza},
title = {tsai - A state-of-the-art deep learning library for time series and sequential data},
howpublished = {Github},
year = {2023}
}

@misc{do2,
author = {Keinert, M. and Schindler-Gmelch, L. and Rupp, L. H. and Sadeghi, M. and Capito, K. and Hager, M. and Fahimi, F. and Richer, R. and Egger, B. and Eskofier, B. M. and Berking, M.},
title = {Facing Depression: Evaluating the Efficacy of the EmpkinS-EKSpression Reappraisal Training augmented with Facial Expressions – Protocol of a Randomized Controlled Trial},
howpublished = {Manuscript submitted for publication},
year = {2024}
}

@article{sil_score,
author = {Rousseeuw, J.},
title = {Silhouettes: a graphical aid to the interpretation and validation of cluster analysis},
journal = {J Comput Appl Math},
year = {1987},
volume = {20},
pages = {53--65}
}

@article{rocket,
author = {Dempster, A. and Schmidt, D. F. and Webb, G. I.},
title = {ROCKET: Exceptionally fast and accurate time series classification using random convolutional kernels},
journal = {Data Mining and Knowledge Discovery},
year = {2019},
volume = {33},
number = {6},
pages = {2066--2095}
}

@article{inception,
author = {Fawaz, H. I. and Forestier, G. and Weber, J. and Idoumghar, L. and Muller, P. A.},
title = {InceptionTime: Finding AlexNet for time series classification},
journal = {arXiv preprint arXiv:1909.04939},
year = {2019}
}

@article{lstm,
author = {Hochreiter, S. and Schmidhuber, J.},
title = {Long short-term memory},
journal = {Neural Computation},
year = {1997},
volume = {9},
number = {8},
pages = {1735--1780}
}

@inproceedings{xgb,
author = {Chen, T. and Guestrin, C.},
title = {XGBoost: A scalable tree boosting system},
booktitle = {Proceedings of the 22nd ACM SIGKDD International Conference on Knowledge Discovery and Data Mining (KDD '16)},
year = {2016},
pages = {785--794}
}

\end{document}